\documentclass[10pt,twocolumn,letterpaper]{article}
\usepackage[cvprfinal]{cvpr}      
\usepackage{amsmath}
\usepackage{times}
\usepackage{graphicx}
\usepackage{enumitem}
\usepackage{subfigure}
\usepackage{amssymb}
\usepackage{multirow}
\usepackage{makecell} 
\usepackage{pifont}
\newcommand{\cmark}{\ding{51}}%
\newcommand{\xmark}{\ding{55}}%

\usepackage[pagebackref,breaklinks,colorlinks]{hyperref}

\def\cvprPaperID{55} 
\def\confName{3DV\xspace}
\def\confYear{2024\xspace}

\begin{document}

\title{LocPoseNet: Robust Location Prior for Unseen Object Pose Estimation}

\author{
Chen Zhao$^{1}$ \quad Yinlin Hu$^3$ \quad Mathieu Salzmann$^{12}$ \\
$^1$EPFL-CVLab \quad $^2$ClearSpace SA \quad $^3$Magic Leap  \\
{\tt\small \{chen.zhao, mathieu.salzmann\}@epfl.ch} 
{\tt\small \{huyinlin\}@gmail.com}
}

\maketitle

\begin{abstract}
Object location prior is critical for the standard 6D object pose estimation setting. The prior can be used to initialize the 3D object translation and facilitate 3D object rotation estimation. Unfortunately, the object detectors that are used for this purpose do not generalize to unseen objects. Therefore, existing 6D pose estimation methods for unseen objects either assume the ground-truth object location to be known or yield inaccurate results when it is unavailable. In this paper, we address this problem by developing a method, LocPoseNet, able to robustly learn location prior for unseen objects. Our method builds upon a template matching strategy, where we propose to distribute the reference kernels and convolve them with a query to efficiently compute multi-scale correlations. We then introduce a novel translation estimator, which decouples scale-aware and scale-robust features to predict different object location parameters. Our method outperforms existing works by a large margin on LINEMOD and GenMOP. We further construct a challenging synthetic dataset, which allows us to highlight the better robustness of our method to various noise sources. Our project website is at: \href{https://sailor-z.github.io/projects/3DV2024_LocPoseNet.html}{https://sailor-z.github.io/projects/3DV2024\_LocPoseNet.html}.
\end{abstract}

\section{Introduction}
\label{sec:intro}
\begin{figure}[!t]
	\centering
    \includegraphics[width=1.0\linewidth]{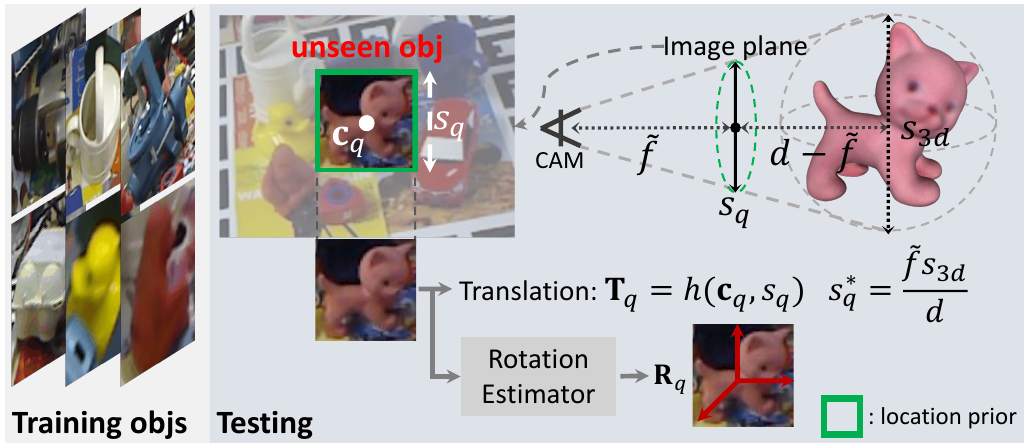}
	\caption{\textbf{Location prior for unseen object pose estimation.} We tackle the case where the unseen object comes from a category (cat) that was not included in the training data. An accurate location prior greatly facilitates 6D object pose estimation: Under a pinhole camera model, it provides valuable information about object translation. Furthermore, it facilitates rotation prediction by cropping the object, thus limiting the influence of the background.} 
    \label{fig:intro}
\end{figure}

6D object pose estimation is a fundamental step in many computer vision and robotics tasks, such as augmented reality~\cite{azuma1997survey}, autonomous driving~\cite{geiger2012we,chen2017multi,xu2018pointfusion,marchand2015pose}, and robotic manipulation~\cite{collet2011moped,zhu2014single,tremblay2018deep}. The state-of-the-art methods~\cite{wang2021gdr,su2022zebrapose} typically follow a two-stage approach: They first locate the object of interest using standard detectors~\cite{ren2015faster,redmon2016you,he2017mask}, and then estimate the object pose from the corresponding cropped image patch. As illustrated in Fig.~\ref{fig:intro}, the object location prior greatly facilitates the pose estimation process.

Unfortunately, these methods are designed to handle the scenario where the objects seen at test time are the same as the training ones. By contrast, in real-world applications, novel objects (unseen during training) are ubiquitous. We have therefore witnessed an increasing number of methods designed to generalize to such novel objects without re-training or fine-tuning~\cite{zhao2022fusing,liu2022gen6d,shugurov2022osop,sun2022onepose,park2020latentfusion}. These methods, however, cannot utilize the same localization strategy as the instance-specific techniques~\cite{wang2021gdr,su2022zebrapose} discussed above, because the widely-used detectors cannot handle new objects from unseen categories. In such context, some approaches~\cite{zhao2022fusing,park2020latentfusion,he2022fs6d} bypass the localization problem by assuming the ground-truth object locations to be available even at test time. While others~\cite{liu2022gen6d,shugurov2022osop} overcome this assumption, the 6D object pose accuracy drops dramatically~\cite{liu2022gen6d} when using predicted object locations instead of the ground-truth ones.

\begin{figure}[!t]
	\centering
        \includegraphics[width=1.0\linewidth]{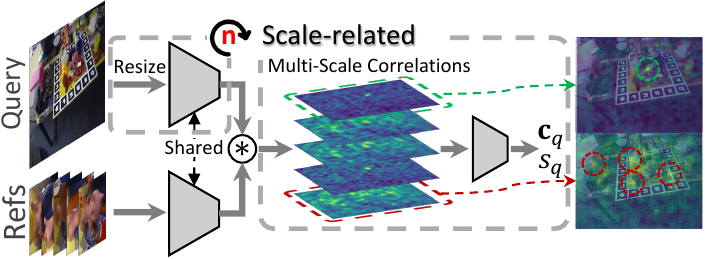}
	\caption{\textbf{Limitations of previous methods.} Multi-scale correlations are computed by passing the query image at $n$ different resolutions through the feature extraction backbone, which is inefficient. Moreover, it yields noisy correlation maps with incorrect high responses (red circles), which in turn interfere with the prediction of the object center $\mathbf{c}_q$.}
	\label{fig:drawback}
\end{figure}

In this paper, we therefore focus on providing accurate location prior for unseen object pose estimation from an RGB image. Our approach is motivated by the following observations. As shown in Fig.~\ref{fig:drawback}, existing methods~\cite{liu2022gen6d,shugurov2022osop,osokin2020os2d,li2018high} handle a novel object by comparing the query image with multiple reference images which depict the object in different orientations. Specifically, the features extracted from the references are convolved with those of the query. To handle the query-reference object size mismatch,
the query image is resized to different $n$ resolutions~\cite{liu2022gen6d}. Unfortunately, this then incurs $n$ forward passes through the feature extraction backbone, which is computationally expensive and time consuming. It also yields noisy correlation maps with incorrect high responses (red circles in Fig.~\ref{fig:drawback}). As the object location parameters are predicted by combining all correlations maps, the prediction tends to be affected by the noise.

In this paper, we therefore develop a method, LocPoseNet, that robustly and efficiently produces a location prior for previously-unseen object pose estimation. To this end, we reason about the individual properties of object location parameters, i.e., a 2D object center and a scalar for object size. Specifically, we explicitly separate the scale-related features, such as the multi-scale correlations discussed above, into scale-robust and scale-aware components. We then predict the object center and the size from the scale-robust features and scale-aware ones, respectively, further leveraging the consistencies across neighboring references. Moreover, we introduce a computationally-friendly approach to extract the scale-related features, which bypasses the need for multiple forward passes through the backbone. This is achieved by leveraging the fact that the receptive field of the reference-query convolution is related to the size of a reference kernel. Thus, we estimate multi-scale correlations by distributing the reference kernels in different manners in the convolution process.

Our experiments on LINEMOD~\cite{hinterstoisser2012model}, GenMOP~\cite{liu2022gen6d}, and a challenging synthetic dataset evidence that our approach yields considerably better localization accuracy than previous methods for novel objects. We then perform 6D object pose estimation by combining our predicted location prior with an off-the-shelf 3D rotation estimator~\cite{liu2022gen6d}. Ours results demonstrate that our object location prior yields a significant boost in unseen object 6D pose estimation accuracy. Our approach also shows better robustness to phenomena such as varying backgrounds, illuminations, object scales, and the reference-query domain gap. 

Our contributions can be summarized as follows:
\begin{itemize}
    \item We introduce a method that provides a robust location prior for unseen objects, which facilitates 6D object pose estimation. 
    \item We develop a decoupled estimator that separates scale-related features and exploits cross-reference consistencies.
    \item We propose to efficiently estimate multi-scale correlations by distributing the reference kernels during the reference-query convolution.
\end{itemize}
Our code and synthetic data will be made publicly available.


\section{Related Work}
\label{sec:related}
\begin{figure*}[!t]
	\centering
	\includegraphics[width=0.9\linewidth]{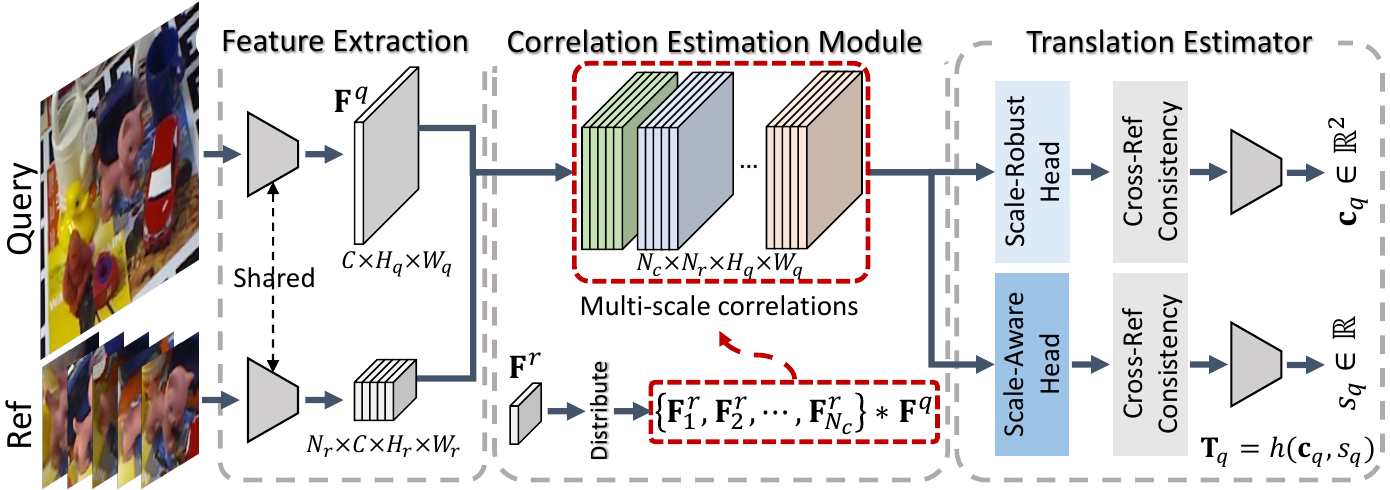}
	\caption{\textbf{Network architecture.} Our network takes a query and a set of references as input. The feature extraction backbone is shared by the query and references. We efficiently capture multi-scale correlations with adjustable receptive fields over the query image (indicated by different colors). The correlations are fed into the presented estimator, where the scale-robust and scale-aware features are separately learned. We predict the object location parameters utilizing cross-reference consistencies, and then compute 3D object translation by using Eq.~\ref{eq:translation}.}
	\label{fig:framework}
\end{figure*}

\noindent{\textbf{Instance-specific 6D pose estimation.}} 6D object pose estimation has been widely studied in the literature. Most existing deep learning methods are instance-specific~\cite{xiang2017posecnn,peng2019pvnet,wang2019densefusion,wang2021gdr}, meaning that the training and testing data contain the same object instances. Although each object instance is observed under different poses during training and testing, the multi-view images in the training set make the deep networks able to memorize the object patterns. This, however, conflicts with the needs of many real-world applications, where objects not included in the training data inevitably exist. The appearance of such novel objects could significantly differ from that of the training ones, which makes instance-specific approaches ill-suited to handle this scenario. 
\\

\noindent{\textbf{Generalizable 6D object pose estimation.}} Recently, efforts have been made towards unseen object pose estimation~\cite{liu2022gen6d,wang2019normalized}. In practice, the unseen objects can be either new instances from seen categories, or objects from entirely novel categories. In the former case, existing approaches~\cite{wang2019normalized,li2020category,chen2020learning} estimate normalized object coordinate space (NOCS) maps from \emph{RGBD} images, and compute the 6D object pose from 3D-3D correspondences. As the training and testing data contain the same object categories, off-the-shelf detection methods~\cite{he2017mask,redmon2016you,liu2016ssd} can be employed to locate the objects. However, such category-level approaches struggle in the presence of objects from new categories. Moreover, the required depth information is not always available, making the RGBD-based methods restricted to specific scenarios. 
This has therefore given rise to   
solutions that handle objects from new categories using \emph{RGB} images. The vast majority~\cite{zhao2022fusing,park2020latentfusion,he2022fs6d,xiao2019pose,sun2022onepose} assumes that the ground-truth object location is available even at test time. Nevertheless, a few approaches~\cite{shugurov2022osop,liu2022gen6d}, which build upon a template matching strategy, can handle unknown object locations. However, the resulting 6D object pose estimates suffer from the inaccurate localization of unseen objects. Developing a method capable of accurately locating novel objects in \emph{RGB} images is therefore pivotal to generalizable 6D object pose estimation.
\\

\noindent{\textbf{One-shot object detection.}} Most object detection methods~\cite{he2017mask,redmon2016you,ren2015faster} follow a category-level learning pipeline, which limits their generalization ability to objects from new categories. To address this issue, some methods~\cite{li2018high,osokin2020os2d,chen2021adaptive,zhao2022semantic,wang2019fast} utilize a template matching strategy. Specifically, they assume the availability of reference images that contain the novel object as templates. Given a query image, correlations between the query and references are computed, and the object location parameters are predicted from the correlation maps. This template matching paradigm has shown promising generalization ability to unseen objects~\cite{liu2022gen6d}. However, it does not account for the specific properties of different location parameters, i.e., object center and size. This is what we achieve in this paper, for the purpose of 6D pose estimation.

\section{Method}
\label{sec:method}
\subsection{Problem Formulation}
\label{sec:formulate}
Given an RGB query image $\textbf{I}_{q}$, we aim to locate an object $O$, with the goal of later using the object location information for 6D object pose estimation. 
We encode such an object location prior with a 2D object center $\textbf{c}_{q}$ and a size scalar $s_{q}$. As shown in Fig.~\ref{fig:intro}, in contrast to the object detection domain~\cite{he2017mask,liu2016ssd,redmon2016you}, the ground-truth $s_{q}^{*}$ is defined based on a pinhole camera model~\cite{hartley2003multiple} as $s_q^{*}=\tilde{f}s_{3d}/d$, where $\tilde{f}$ is a virtual focal length~\cite{liu2022gen6d}, $s_{3d}$ denotes the size of the 3D object model, and $d$ is the depth of the object center. In this paper, we focus on locating unseen objects, which means that the objects observed during testing are not part of the training data, i.e., $\mathcal{O}_{train}\cap \mathcal{O}_{test}=\emptyset$. Following~\cite{liu2022gen6d,shugurov2022osop}, for each $O\in\mathcal{O}_{train}\cup\mathcal{O}_{test}$, we assume to have access to a set of reference images $\mathcal{I}_r=\{\textbf{I}_{r}^{1}, \textbf{I}_{r}^{2},\dots, \textbf{I}_{r}^{N_{r}}\}$ which depict $O$ from different viewpoints. The object location is known for every $\textbf{I}_{r}^{i}, i=1,2,\dots,N_r$. The references are resized to the same resolution, and therefore all object sizes $\{s_{r}^{1},s_{r}^{2},\dots,s_{r}^{N_r}\}$ are equal. 

\subsection{Framework Overview}
Fig.~\ref{fig:framework} depicts the overall architecture of our network. As template matching has been shown to be effective to handle unseen objects~\cite{liu2022gen6d}, we follow such a paradigm, using a Siamese network to extract features from the query and reference. The reference feature map then acts as a convolution kernel over the query ones. As will be discussed in more detail below, to account for the potential object size mismatch between the query and reference, we estimate multi-scale correlations in a more efficient and scalable manner. We then explicitly separate this multi-scale information into scale-robust ans scale-aware components, which are exploited to predict the respective object location parameters, while leveraging consistencies across multiple references.

\subsection{Efficient Multi-Scale Correlation Estimation}
\label{sec:correlation}
Typically, the object sizes in the query and the reference are different. In~\cite{liu2022gen6d}, this problem is tackled by resizing the query to $n$ different resolutions and feeding all resized images to the network. While this indeed captures multi-scale correlations, it comes at the computation cost of performing $n$ forward passes through the feature extraction backbone. Here, instead, we propose to distribute a reference kernel at different spatial resolutions to obtain multi-scale correlations in a more efficient and scalable manner.

\begin{figure}[!t]
	\centering
	\includegraphics[width=1.0\linewidth]{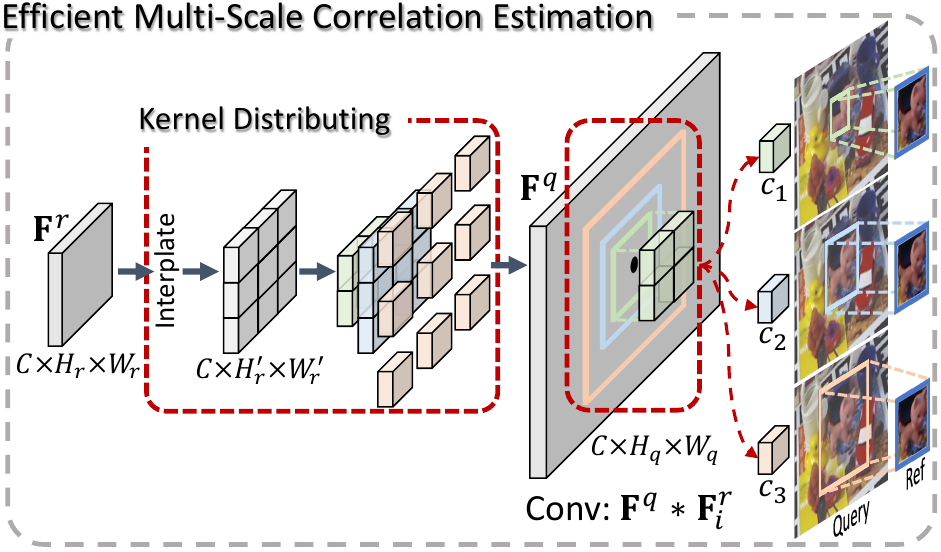}
	\caption{\textbf{Efficient multi-scale correlation estimation module.} A reference kernel is distributed at different spatial sizes. The query feature map $\mathbf{F}^{q}$ is then convolved with all the ditributed kernels. The resulting $(c_1,c_2,c_3)$ capture information from the query with different receptive fields, as indicated by the colored boxes.}
	\label{fig:conv}
\end{figure}

Formally, let us express the reference-query convolution as
\begin{align}
\label{eq:conv}
f(\textbf{F}^{c}, \textbf{F}^{r})=\sum_{i=1}^{H_r}\sum_{j=1}^{W_r}\textbf{F}_{ij}^{c}\cdot\textbf{F}_{ij}^{r},
\end{align}
where $\textbf{F}^{r}\in\mathbb{R}^{C\times H_r\times W_r}$ is a reference kernel, $\textbf{F}^{c}\in\mathbb{R}^{C\times H_r\times W_r}$ indicates the region in the query feature map covered by the kernel at each spatial location during the convolution, and ${\bf u} \cdot {\bf v}$ denotes a dot product. In this simple formulation, the receptive field over the query feature map has exactly the same size as the reference kernel itself. To obtain convolutions with an adjustable receptive field, which thus yields multi-scale correlations, we propose to distribute the reference kernel over different spatial resolutions, as shown in Fig.~\ref{fig:conv}. 

Specifically, we first sample a set of offsets $\{\Delta{o_i}, \;1\leq i \leq N\}$. For each offset, we then interpolate a reference kernel at the corresponding resolution. This yields a set of candidate kernels $\mathcal{F}^{r}=\{\mathbf{F}_1^{r}, \mathbf{F}_2^{r},\dots,\mathbf{F}_N^{r}\}, \mathbf{F}_i^{r}\in\mathbb{R}^{C\times (H_r+\Delta{o_i})\times(W_r+\Delta{o_i})}$.
To account for large scale variations, we further perform a \emph{bidirectional} process over the candidates. We shrink and expand a candidate by utilizing pyramid pooling~\cite{he2015spatial} and dilation~\cite{chen2017rethinking}, respectively. In these two cases, the size of a distributed kernel is computed as
\begin{align}
\label{eq:pool}
H_r^{'}=\left \lceil \frac{H_r+\Delta{o_i}}{t} \right \rceil,
\end{align}
and 
\begin{align}
\label{eq:dilated}
H_r^{'}=(H_r+\Delta{o_i})t-t+1,
\end{align}
respectively, where $t$ denotes the rate of pooling and dilation, and we take $H_r$ as an example but apply the same process to $W_r$. We then stack the correlations as 
\begin{align}
\label{eq:hybrid}
\mathbf{C} = \text{CAT}[(\textbf{F}^{q}*\textbf{F}_{1}^{r}), (\textbf{F}^{q}*\textbf{F}_{2}^{r}),\dots,(\textbf{F}^{q}*\textbf{F}_{N_{c}}^{r})],
\end{align}
where $\textbf{C}\in\mathbb{R}^{N_{c}\times{H_q}\times{W_q}}$, \emph{CAT} represents concatenation, $*$ denotes convolution, and $N_c$ is the number of distributed kernels. Benefiting from this two-stage strategy, our method is capable of handling both fine-grained and large-scale size mismatches. Importantly, distributing the reference kernel only requires a single pass through the feature extraction backbone, which makes our method more efficient and scalable than~\cite{liu2022gen6d}.

Notably, the \emph{bidirectional} kernel distribution is crucial in our scenario where the query object could be larger or smaller than the reference one. Any \emph{unidirectional} process~\cite{peng2021attentional,qiao2021detectors} may further worsen the object size mismatch. For example, the dilation is used to expand the receptive field over the query ($t > 1$); it would aggravate the mismatch when the query object is smaller than the reference. 
\begin{figure}[!t]
	\centering
	\subfigure[Raw Correlations]
	{\includegraphics[height=2.25cm]{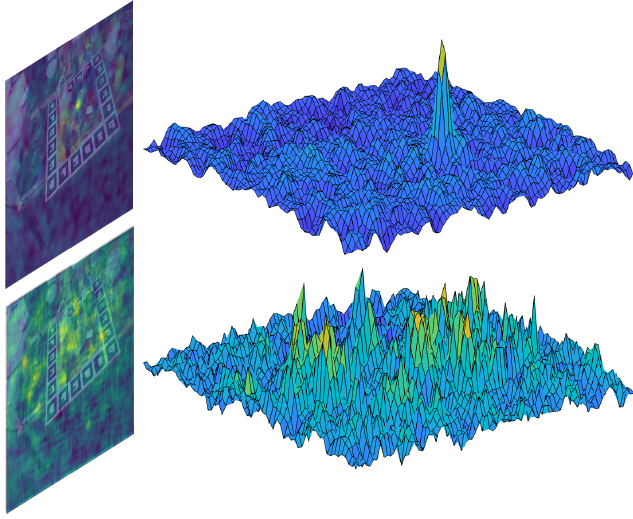}\label{fig:corr_init}
	}
	\subfigure[Deterministic Feature Fusion]
	{\includegraphics[height=1.95cm]{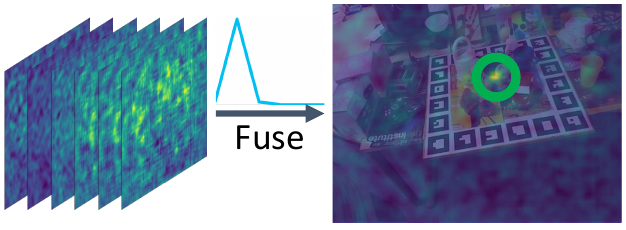}\label{fig:corr_fused}
	}
        \caption{\textbf{Illustrations of the correlation maps and the deterministic feature fusion.} (a) shows two correlation maps and the bottom one with multiple peaks is unreliable when computing the object center. To mitigate the impact of such noisy correlation maps, we present a fusion approach where smaller weights are assigned to the noisy maps in a deterministic manner. The blue curve represents the computed weights.}
	\label{fig:viz_corr}
\end{figure}

\subsection{Decoupled Translation Estimator}
As introduced in Fig.~\ref{fig:intro}, the object translation can be computed from $(\mathbf{c}_q, s_q)$. Therefore, effectively predicting $(\mathbf{c}_q, s_q)$ from $\textbf{C}\in\mathbb{R}^{N_{c}\times{H_q}\times{W_q}}$ is pivotal to high-accuracy translation estimation. The previous method~\cite{liu2022gen6d} applies several convolutional layers over $\textbf{C}$, which outputs the predictions of $\mathbf{c}_q$ and $s_q$. Recall that $\textbf{C}$ represents multiple correlation maps that contain multi-scale features. We argue that the predictions of the two location parameters have conflicting requirements w.r.t. the scale-related features, and propose to estimate them in a decoupled manner.

Intuitively, we expect the object center to correspond to the maximum response in each of the correlation maps. However, as the correlation maps correspond to different receptive fields over the query image, those with ill-suited receptive fields could be noisy and unreliable, as illustrated in Fig.~\ref{fig:corr_init}. To alleviate this issue, we propose to feed the raw correlations to a scale-robust head for object center prediction, which fuses the $N_c$ correlation maps in a deterministic manner. More specifically, we measure the reliability of a correlation map based on the distribution of the correlation scores. As shown in Fig.~\ref{fig:corr_init}, we expect a distribution with one peak (top one) to be more reliable than the one with multiple peaks (bottom one). Taking computational stability into account, we first normalize each correlation map as
\begin{align}
\label{eq:normal}
\tilde{\mathbf{C}}_i=\frac{\mathbf{C}_i-\mu_i}{\sigma_i},
\end{align}
where $\mathbf{C}_i\in\mathbb{R}^{{H_q}\times{W_q}}$ is the $i$-th correlation map, $\mu_i$ denotes its mean, and $\sigma_i$ indicates its standard deviation.
We then compute a weight for $\tilde{\mathbf{C}}_i$ as 
\begin{align}
\label{eq:weight}
w_i=\frac{1}{H_qW_q}\sum_{n=1}^{H_q}\sum_{m=1}^{W_q}{\text{max}(\tilde{\mathbf{C}}_i)-\tilde{c}_{nm}},
\end{align}
where $\tilde{c}_{nm}\in\mathbb{R}$ is an element of $\tilde{\mathbf{C}}_i$. Subsequently, we fuse all correlation maps as  
\begin{align}
\label{eq:fuse}
\tilde{\mathbf{C}}=\sum_{i=1}^{N_c}{\frac{\text{exp}(w_i)\tilde{\mathbf{C}}_i}{\sum_{i=1}^{N_c}{\text{exp}(w_i)}}}, \ \ \tilde{\mathbf{C}}\in\mathbb{R}^{H_q\times{W_q}}.
\end{align}
Fig.~\ref{fig:corr_fused} demonstrates the effectiveness of the proposed deterministic fusion approach, where the blue curve represents the computed weights. A dominating weight is assigned to the most reliable correlation map, which makes the fused result more reliable than $\mathbf{C}$.

By contrast, to predict the object size, one needs to reason about the object patterns at different scales. Such information is contained in the initial multi-scale correlations $\mathbf{C}$. However, this property would be lost after the fusion process in the scale-robust head. Therefore, we introduce a scale-aware head that exploits the full size-related information available in $\mathbf{C}$ by extracting an embedding $\hat{\mathbf{C}}=f(\mathbf{C}, \Theta)$, where $\hat{\mathbf{C}}\in\mathbb{R}^{D\times{H_q}\times W_q}$, $f$ denotes fully-connected layers with learnable parameters $\Theta$.

Note that, so far, we have considered the case of a single reference image, but in practice we have access to $N_r$ references. To leverage them and obtain more robust features for object localization, we propose to exploit consistencies across the references. Concretely, we measure the distance between two references based on the geodesic distance of the corresponding object poses, computed as 
\begin{align}
\label{eq:distance}
d_{ij}=\text{arccos}\left(\frac{\text{tr}(\mathbf{R}_{i}^{\text{T}}\mathbf{R}_{j})-1}{2}\right)/\pi,
\end{align}
where $\mathbf{R}_{i}$ and $\mathbf{R}_{j}$ represent the object rotations in the $i$-th and $j$-th reference images, respectively. Intuitively, the correlations estimated from neighboring references should be consistent with each other. To account for this, we pick $k$-nearest neighbors for each reference based on Eq.~\ref{eq:distance} and concatenate the corresponding correlations.
We denote the resulting feature maps from the scale-robust and scale-aware heads as $\textbf{M}_{rb}\in\mathbb{R}^{N_r\times{k}\times{H_q}\times{W_q}}$ and $\textbf{M}_{aw}\in\mathbb{R}^{N_r\times{kD}\times{H_q}\times{W_q}}$, respectively. These two feature maps are further separately embedded into a latent space $\mathbb{R}^{D\times{H_q}\times{W_q}}$ after being fed into several convolution blocks and a max pooling layer ($\mathbb{R}^{N_r\times{D}\times{H_q}\times{W_q}}\rightarrow\mathbb{R}^{{D}\times{H_q}\times{W_q}}$).

Subsequently, we separately predict the object center and the object size from the two embedded feature maps. Given the predicted object location parameters $\textbf{c}_q$ and $s_q$, we compute the 3D object translation as
\begin{align}
\label{eq:translation}
\textbf{T}_q=h(\textbf{c}_q,s_q)=\frac{\tilde{f}s_{3d}}{s_q}\textbf{K}^{-1}\hat{\textbf{c}}_q,
\end{align}
where $\textbf{K}$ is the known camera intrinsic matrix, and $\hat{\textbf{c}}_q=[\textbf{c}_q, 1]^{T}$.

\section{Experiments}
\label{sec:exp}
\noindent \textbf{Implementation details.}
Following the implementation in~\cite{liu2022gen6d}, we employ a pre-trained VGG-11~\cite{simonyan2014very} network as the feature extraction module. The detail of our network architecture and kernel distributing configuration is provided in the supplementary material. We train our network for $300,000$ iterations using the Adam optimizer~\cite{kingma2014adam} with a batch size of 8 and a learning rate of $10^{-4}$, divided by 5 after every $100,000$ iterations. We set $k=3$ in the cross-reference consistency learning module by default. When extending to 6D object pose estimation, we combine our method with \cite{liu2022gen6d} to estimate 3D object rotation.
\\

\noindent \textbf{Experimental setup.}
\begin{figure}[!t]
	\centering
	\includegraphics[width=0.9\linewidth]{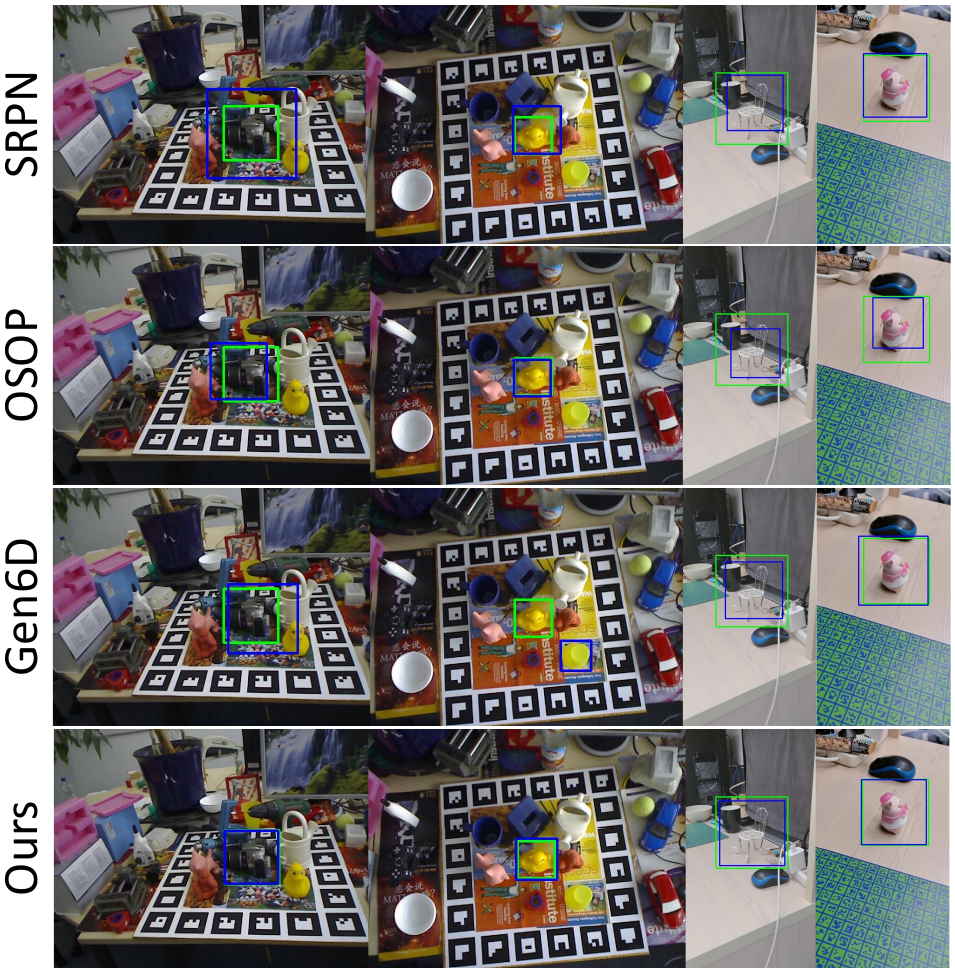}
	\caption{\textbf{Qualitative results on LINEMOD~\cite{hinterstoisser2012model} and GenMOP~\cite{liu2022gen6d}.} The green and blue object bounding boxes represent the ground-truth and predicted ones, respectively. }
	\label{fig:real_res}
\end{figure}
We conduct experiments on two public datasets, i.e., LINEMOD~\cite{hinterstoisser2012model} and GenMOP~\cite{liu2022gen6d}. To achieve a fair comparison, we follow Gen6D's~\cite{liu2022gen6d} settings and train our network using the same training data as Gen6D. Note that all objects in the testing phase are previously unseen, i.e., not included in the training data. We compare our method with existing ones from three perspectives, i.e., unseen object localization, robustness, and unseen object 6D pose estimation.
\subsection{Unseen Object Localization}
\begin{table}[!t]
	\begin{center}
	\small
		\begin{tabular}{m{1.72cm}m{0.6cm}<{\centering}m{0.6cm}<{\centering}m{0.6cm}<{\centering}m{0.6cm}<{\centering}m{0.7cm}<{\centering}m{0.6cm}<{\centering}}
		    \Xhline{2\arrayrulewidth}
		   & Cat & Duck & Bvise & Cam & Driller & \textbf{Avg.} \\
		    \hline
    	   SRPN~\cite{li2018high} & 9.72 & 4.56 & 22.47 & 13.43 & 10.97 & 12.23 \\
    	   SRPN-P & 11.85 & 1.62 & 18.94 & 2.44 & 8.91 & 8.76 \\
    	   SRPN-D & 22.97 & 1.85 & 49.14 & 17.76 & 18.89 & 22.12 \\
		   OSOP~\cite{shugurov2022osop} & 32.10 & 34.81 & 26.68 & 24.33 & 21.36 & 27.86 \\
		   Gen6D~\cite{liu2022gen6d} & 76.99 & 42.15 & 63.33 & 72.92 & 48.78 & 60.84 \\
		   \textbf{Ours} & \textbf{81.68} & \textbf{61.80} & \textbf{79.45} & \textbf{80.50} & \textbf{68.31} & \textbf{74.35} \\
		    \Xhline{2\arrayrulewidth}
			\end{tabular}
	\end{center}
        \caption{\textbf{Unseen object localization on LINEMOD~\cite{hinterstoisser2012model}.} We report the mAP@[.5:.95] (\%)~\cite{lin2014microsoft}.}
	    \label{tab:linemod_T}
\end{table}

\begin{table}[!t]
	\begin{center}
	\small
		\begin{tabular}{m{1.72cm}m{0.6cm}<{\centering}m{0.8cm}<{\centering}m{0.6cm}<{\centering}m{0.55cm}<{\centering}m{0.55cm}<{\centering}m{0.6cm}<{\centering}}
		    \Xhline{2\arrayrulewidth}
		   & Chair & PlugEN & Piggy & Sci. & TF. & \textbf{Avg.} \\
		    \hline
		   SRPN~\cite{li2018high} & 7.17 & 69.16 & 51.46 & 12.92 & 57.76 & 39.69 \\
		   SRPN-P & 8.64 & 81.68 & 60.29 & 9.44 & 30.32 & 38.07 \\
		   SRPN-D & 34.39 & 79.13 & 84.84 & 35.65 & 65.71 & 59.94 \\
		   OSOP~\cite{shugurov2022osop} & 22.50 & 49.05 & 38.75 & 26.30 & 59.47 & 39.21 \\
		   Gen6D~\cite{liu2022gen6d} & 67.48 & 81.03 & 84.17 & 76.53 & 76.31 & 77.10 \\
		   \textbf{Ours} & \textbf{77.35} & \textbf{83.42} & \textbf{88.76} & \textbf{77.72} & \textbf{87.52} & \textbf{82.95} \\
		\Xhline{2\arrayrulewidth}
		\end{tabular}
	\end{center}
        \caption{\textbf{Unseen object localization on GenMOP~\cite{liu2022gen6d}.} We use the mAP@[.5:.95] (\%)~\cite{lin2014microsoft} as metric.}
	\label{tab:GenMOP_T}
\end{table}
Table~\ref{tab:linemod_T} and Table~\ref{tab:GenMOP_T} provide the mAP@[.5:.95] (\%)~\cite{lin2014microsoft} of the evaluated methods on LINEMOD and GenMOP, respectively. We compare our approach to Gen6D~\cite{liu2022gen6d}, which, as mentioned before, is the most related work to ours. We also report the results of OSOP~\cite{shugurov2022osop}, which is an advanced version of one-shot object detection methods, such as OS2D~\cite{osokin2020os2d}, and achieves better performance as shown in~\cite{shugurov2022osop}. Additionally, we evaluate a representative visual tracking method, SRPN~\cite{li2018high}, since it also follows a template matching paradigm. We employ the pre-trained model provided by the authors for Gen6D. We train OSOP and SRPN from scratch, following the same settings as for Gen6D and our approach. We sample the same 32 reference images for all evaluated methods, which leads to a fair comparison. As shown in Table~\ref{tab:linemod_T} and Table~\ref{tab:GenMOP_T}, our approach achieves significantly better mAPs in all cases, i.e., $\mathbf{13.51\%}$ increase on LINEMOD and $\mathbf{5.85\%}$ increase on GenMOP on average. Moreover, we extend SRPN to a multi-scale version by adding pyramid pooling layers (SRPN-P) or dilated convolutions (SRPN-D). It is worth noting that SRPN-D achieves better performance than the baseline SRPN, whereas SRPN-P leads to a decrease in mAPs. Analyzing the relative object size between the query and reference in LINEMOD and GenMOP revealed that the query objects are typically larger than the reference ones. The results therefore confirm our intuition that a \emph{unidirectional} kernel resizing may aggravate the object size mismatch, as discussed in Section~\ref{sec:correlation}.

Fig.~\ref{fig:real_res} depicts some qualitative results. As explained in Section~\ref{sec:formulate}, the definition of the object location parameters differs from the one in the object detection domain~\cite{he2017mask,liu2016ssd,redmon2016you}. Therefore, all visualized object bounding boxes are square. One can easily observe that our predictions are consistently closer to the ground-truth ones. 

\subsection{Robustness Evaluation}
\begin{table}[!t]
	\begin{center}
	\small
        \begin{tabular}{m{1.72cm}m{0.6cm}<{\centering}m{0.6cm}<{\centering}m{0.6cm}<{\centering}m{0.6cm}<{\centering}m{0.7cm}<{\centering}m{0.6cm}<{\centering}}
            \Xhline{2\arrayrulewidth}
           & Cat & Duck & Bvise & Cam & Driller & \textbf{Avg.} \\
            \hline
           SRPN~\cite{li2018high} & 39.55 & 31.42 & 23.73 & 42.61 & 28.44 & 31.42 \\
           SRPN-P & 37.08 & 13.22 & 32.58 & 22.40 & 28.44 & 26.75 \\
           SRPN-D & 44.50 & 14.13 & 22.53 & 16.84 & 32.68 & 26.14 \\
           OSOP~\cite{shugurov2022osop} & 29.90 & 28.14 & 7.59 & 13.02 & 9.55 & 17.64 \\
           Gen6D~\cite{liu2022gen6d} & 47.68 & 37.83 & 31.90 & 40.49 & 27.98 & 37.18 \\
           \textbf{Ours} & \textbf{56.77} & \textbf{45.01} & \textbf{45.20} & \textbf{51.22} & \textbf{39.82} & \textbf{47.60} \\
            
            \Xhline{2\arrayrulewidth}
            \end{tabular}
	\end{center}
        \caption{\textbf{mAP@[.5:.95] (\%)~\cite{lin2014microsoft} of unseen object locating on synthetic LINEMOD.}}
	\label{tab:robust}
\end{table}

\begin{figure}[!t]
	\centering
	\includegraphics[width=0.9\linewidth]{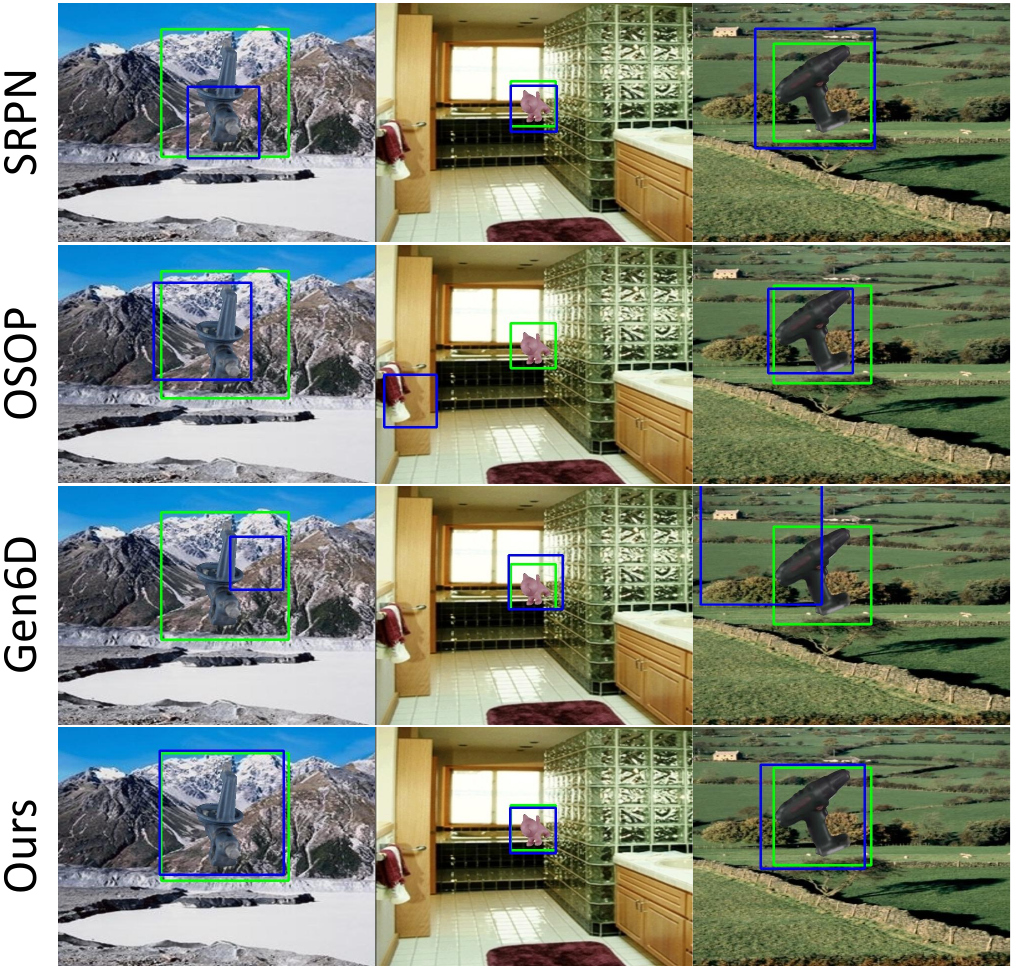}
	\caption{\textbf{Qualitative results on synthetic LINEMOD.}}
	\label{fig:syn_res}
\end{figure}

\begin{figure}[!t]
	\centering
	\subfigure[]
	{\includegraphics[height=2.82cm]{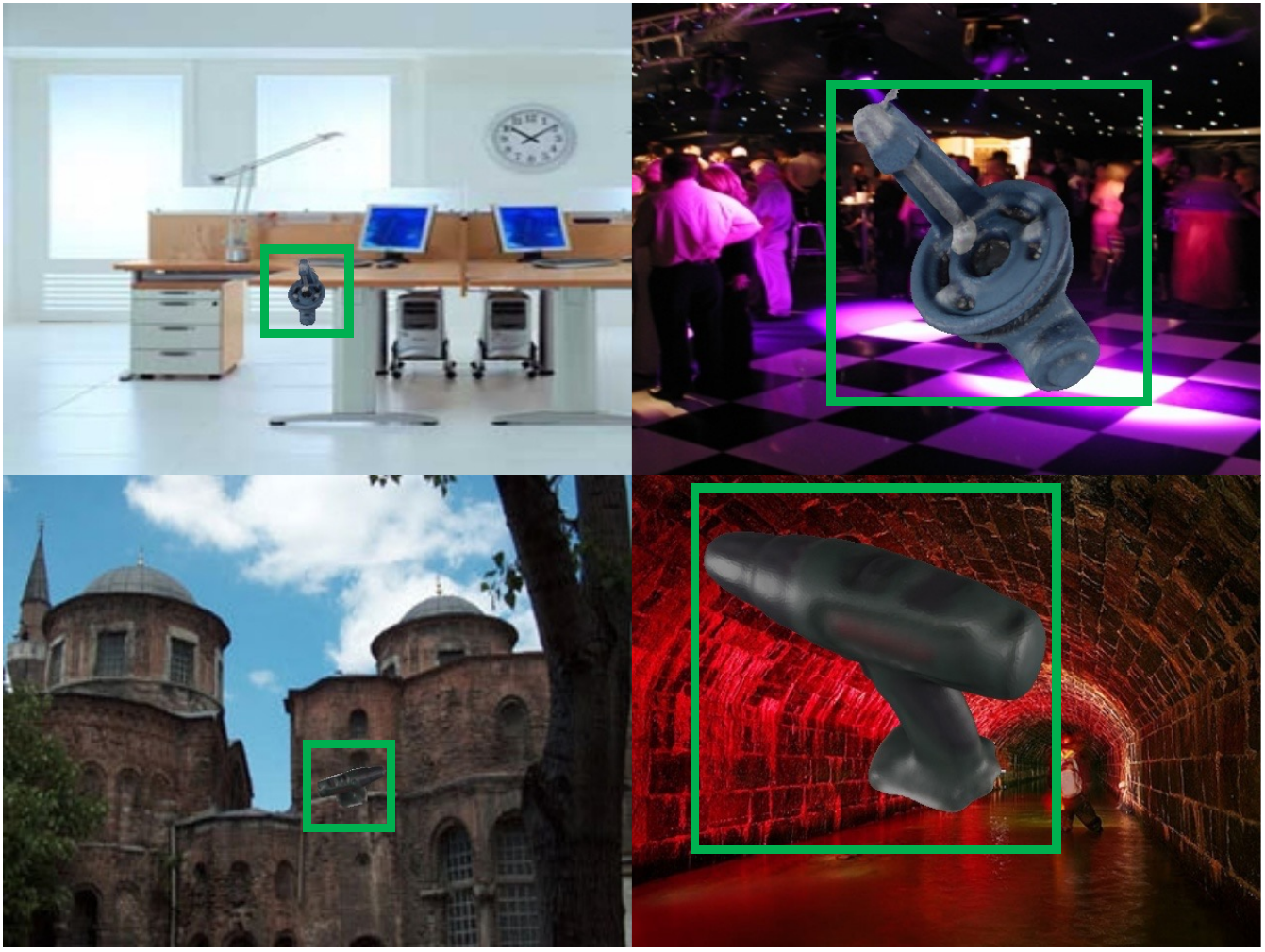}\label{fig:syn_imgs}
	}
	\subfigure[]
	{\includegraphics[height=2.82cm]{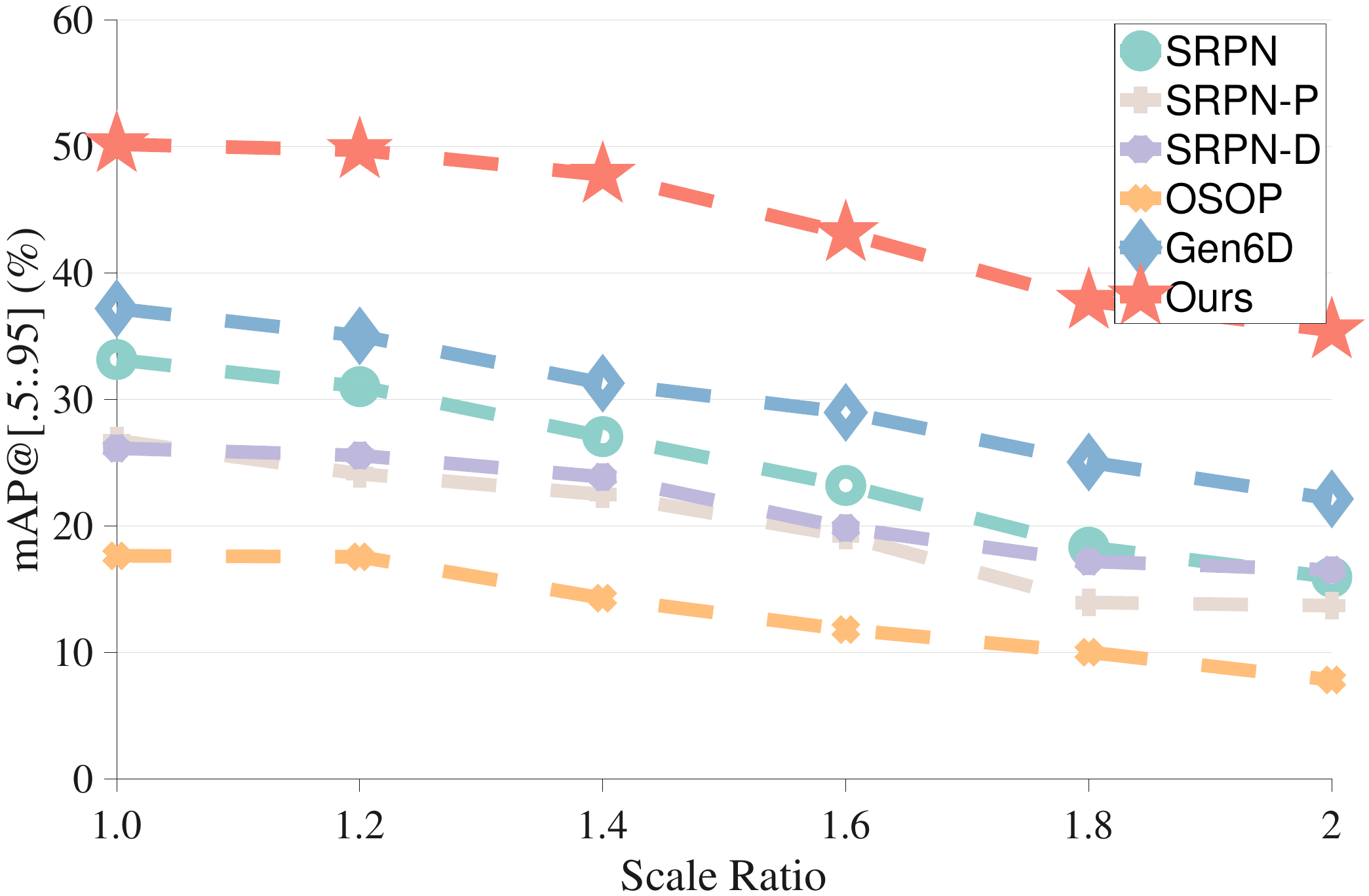}\label{fig:scale_fig}
	}
	\caption{\textbf{Robustness against various noise sources.} (a) Synthetic query images from the constructed dataset, where the object size, illumination, and background widely vary. (b) mAP@[.5:.95] (\%)~\cite{lin2014microsoft} under different scale ratios.}
	\label{fig:scale}
\end{figure}

In Gen6D's experimental setup, the references are real images taken from different viewpoints. The object size and background in the reference images are therefore similar to the ones in the query images. In real applications, however, these factors tend to differ. Moreover, the reference images may often be synthetic ones when the 3D object model is available~\cite{zhao2022fusing,shugurov2022osop}, which leads to a domain gap between the query and references. Taking all these challenges into account, we construct a synthetic dataset to evaluate the robustness of the approaches. Specifically, we keep the original \emph{real} reference images but render the 3D object models of LINEMOD from different viewpoints with diverse illumination conditions to generate \emph{synthetic} query images. We randomly assign a background to each synthetic query using images from SUN2012~\cite{xiao2010sun}. Some exemplar query images are shown in Fig.~\ref{fig:syn_imgs}. We use the combination of real references and synthetic queries here because it enables large-scale object scale mismatches. Note that we only employ this dataset for testing without retraining or fine-tuning. The results on the synthetic dataset are provided in Table~\ref{tab:robust}. The mAPs of all evaluated methods decrease compared with those in Table~\ref{tab:linemod_T}, which demonstrates that our synthetic dataset is more challenging.  Nevertheless, our method still surpasses previous works by a large margin (at least $\mathbf{10.42\%}$). Notably, both SRPN-P and SRPN-D perform worse than SRPN. As the object size mismatch is bidirectional in the synthetic dataset, these unidirectional approaches therefore become ineffective. Fig.~\ref{fig:syn_res} illustrates some visualization results, where our method consistently performs better than the competitors.

To further analyze robustness to object size variations, we resize the objects in the synthetic query images with different scale ratios. This can be formalized as $\hat{s}_q \in [s_{q}/p, s_{q}*p]$, where $p$ represents the scale ratio. We sample 6 scale ratios $p\in\{1.0, 1.2, 1.4, 1.6, 1.8, 2.0\}$ and report the mAP@[.5:.95] in all these cases. The dataset becomes more challenging as $p$ increases. As shown in Fig.~\ref{fig:scale_fig}, the performance of our method is consistently better than that of previous works when the scale ratio varies. This observation further demonstrates the better robustness of our method to various noise sources.

\subsection{6D Unseen Object Pose Estimation}
\begin{table}[!t]
	\begin{center}
		\begin{tabular}{lccc}
	    \Xhline{2\arrayrulewidth}
	    Method & Retrain & Refine & Avg. \\
	     \hline 
	     AAE~\cite{sundermeyer2018implicit} & \cmark & \xmark & 19.63 \\
	     Self6D~\cite{wang2020self6d} & \cmark & \xmark & 51.32 \\
	     DPOD-Syn~\cite{zakharov2019dpod} & \cmark & \xmark & 43.21 \\
	     DPOD-Syn~\cite{zakharov2019dpod} & \cmark & \xmark & 59.18 \\
	     PFS~\cite{xiao2019pose} & \cmark & DeepIM~\cite{li2018deepim} & 53.36 \\
	     PVNet~\cite{peng2019pvnet} & \cmark & \xmark & 83.02 \\
	     PoseCNN~\cite{xiang2017posecnn} & \cmark & DeepIM~\cite{li2018deepim} & 89.14 \\
	     DPOD-Real~\cite{zakharov2019dpod} & \cmark & DPOD~\cite{zakharov2019dpod} & 94.86 \\
	     \hline
		 PFS-GT~\cite{xiao2019pose} & \xmark & \xmark & 15.88 \\
		 Gen6D~\cite{liu2022gen6d} & \xmark & \xmark & 17.73 \\
		 Ours & \xmark & \xmark & \textbf{27.27} \\
		 Gen6D~\cite{liu2022gen6d} & \xmark & Volume~\cite{liu2022gen6d} & 62.45 \\
		 Ours & \xmark & Volume~\cite{liu2022gen6d} & \textbf{68.58} \\
		    \Xhline{2\arrayrulewidth}
		\end{tabular}
	\end{center}
        \caption{\textbf{Performance of 6D pose estimation for unseen objects on LINEMOD~\cite{hinterstoisser2012model}.} We use the ADD-0.1d (\%)~\cite{hinterstoisser2012model} as  evaluation metric. ``Retrain" indicates if the method is retrained on the unseen objects.}
	\label{tab:linemod_6d}
\end{table}
\begin{table}[!t]
	\begin{center}
		\begin{tabular}{lccc}
	    \Xhline{2\arrayrulewidth}
	     Method & Retrain & Refine & Avg. \\
	     \hline 
	     PVNet~\cite{peng2019pvnet} & \cmark & \xmark & 38.79 \\
	     RLLG~\cite{cai2020reconstruct} & \cmark & \xmark & 2.71 \\
	     \hline
		 ObjDesc~\cite{wohlhart2015learning} & \xmark & \xmark & 8.55 \\
		 Gen6D~\cite{liu2022gen6d} & \xmark & \xmark & 17.90 \\
		 Ours & \xmark & \xmark & \textbf{23.32} \\
		 Gen6D~\cite{liu2022gen6d} & \xmark & Volume~\cite{liu2022gen6d} & 50.39 \\
		 Ours & \xmark & Volume~\cite{liu2022gen6d} & \textbf{53.61} \\
	    \Xhline{2\arrayrulewidth}
		\end{tabular}
	\end{center}
        \caption{\textbf{ADD-0.1d (\%)~\cite{hinterstoisser2012model} of 6D unseen object pose estimation on GenMOP~\cite{liu2022gen6d}}.}
	\label{tab:genmop_6d}
\end{table}

To perform 6D unseen object pose estimation with our method, we compute the 3D object translation using Eq.~\ref{eq:translation} and feed our predicted location prior $(\mathbf{c}_q, s_q)$ to an off-the-shelf rotation estimator~\cite{liu2022gen6d}. In Table~\ref{tab:linemod_6d} and Table~\ref{tab:genmop_6d}, we report the ADD-0.1d~\cite{hinterstoisser2012model} for all evaluated methods on LINEMOD and GenMOP, respectively. As the instance-specific methods such as PVNet~\cite{peng2019pvnet} cannot generalize to unseen objects, the models are \emph{retrained} on the testing objects. Therefore, we only compare our method with other generalizable competitors that are not retrained. Note that we assume that the depth information is unavailable, so LatentFusion~\cite{park2020latentfusion} and category-level approaches~\cite{wang2019normalized,li2020category,chen2020learning} are not evaluated. We also exclude OSOP~\cite{shugurov2022osop} because it requires $90,000$ reference images for each object during object pose estimation, which becomes intractable for the $3000$ objects in the training set used in~\cite{liu2022gen6d}. Our method yields better ADDs than existing generalizable approaches when no refinement is performed, which truly shows that the 6D object poses obtained with our method are more accurate. When combined with the same refinement approach as~\cite{liu2022gen6d}, our method also performs better, which leads to competitive results compared with instance-specific methods. These observations evidence that 6D object pose estimation can benefit from an accurate object location prior.
\subsection{Ablation Studies}
\begin{table}[!t]
	\begin{center}
		\begin{tabular}{cccl}
	    \Xhline{2\arrayrulewidth}
	    Multi. & Ref. Cons. & Est. & mAP@[.5:.95] (\%) \\
	    \hline 
	      \xmark & \xmark & \xmark & 34.98 \\
	      \xmark & \cmark & \xmark & 38.41 (+3.43) \\
	      ours & \cmark & max & 52.43 (+14.02)\\
            ours & \cmark & conv & 70.10 (+31.69) \\
            ours & \cmark & ours & \textbf{74.35 (+35.94)} \\
            \hline 
            pool & \cmark & ours & 57.86 (-16.49) \\
            dilate & \cmark & ours & 60.77 (-13.58) \\
            resize\_r & \cmark & ours & 69.81 (-4.54) \\
            resize\_q & \cmark & ours & 73.07 (-1.28) \\
	    \Xhline{2\arrayrulewidth}
		\end{tabular}
	\end{center}
        \caption{\textbf{Effectiveness of each component.} We report results on LINEMOD~\cite{hinterstoisser2012model}. ``Multi." means multi-scale correlations, ``Ref. Cons." represents cross-reference consistency, and ``Est.'' indicates the estimator.}
	\label{tab:aba}
\end{table}
\begin{figure}[!t]
	\centering
	\subfigure[]
	{\includegraphics[width=0.48\linewidth]{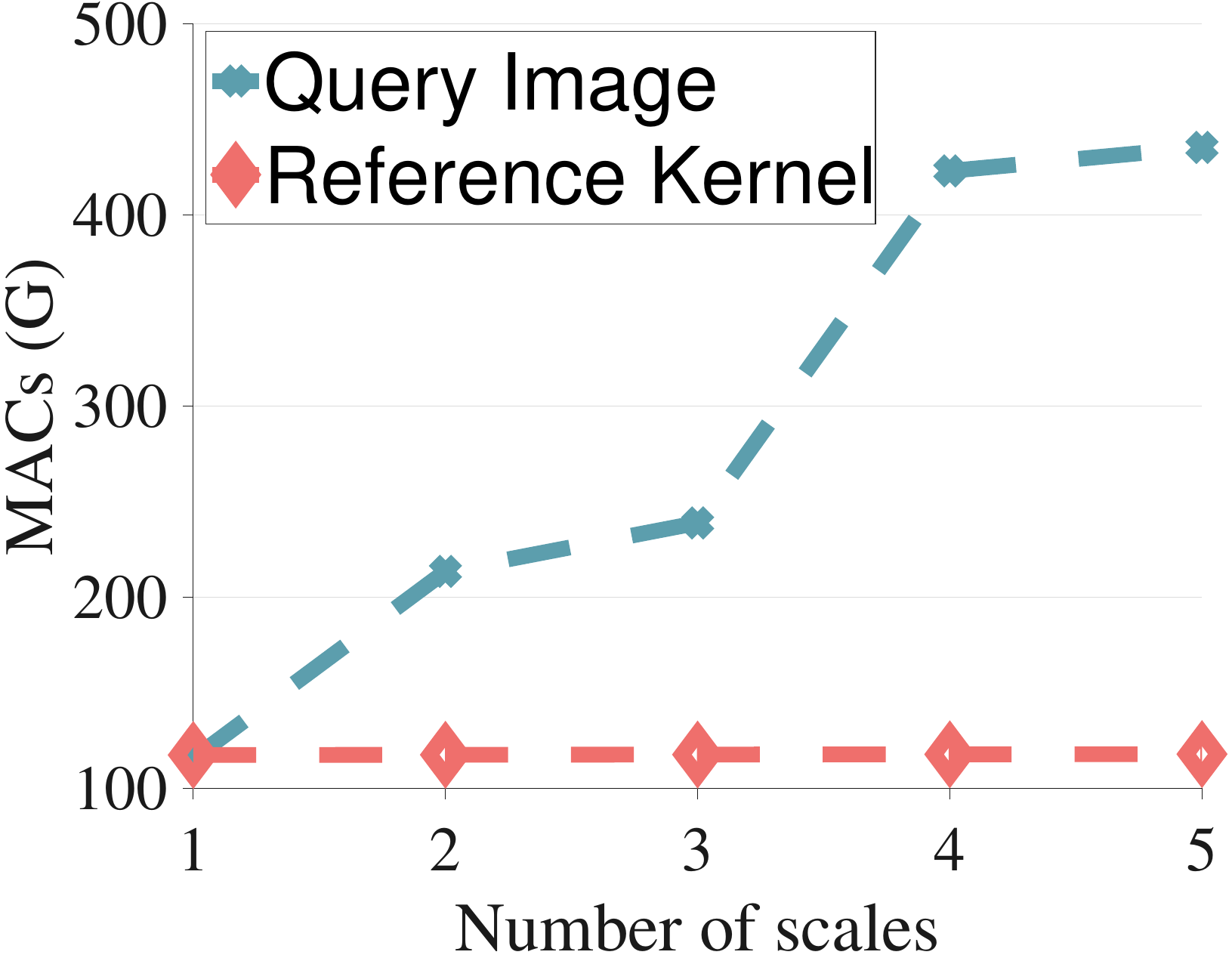}\label{fig:macs}
	}
	\subfigure[]
	{\includegraphics[width=0.48\linewidth]{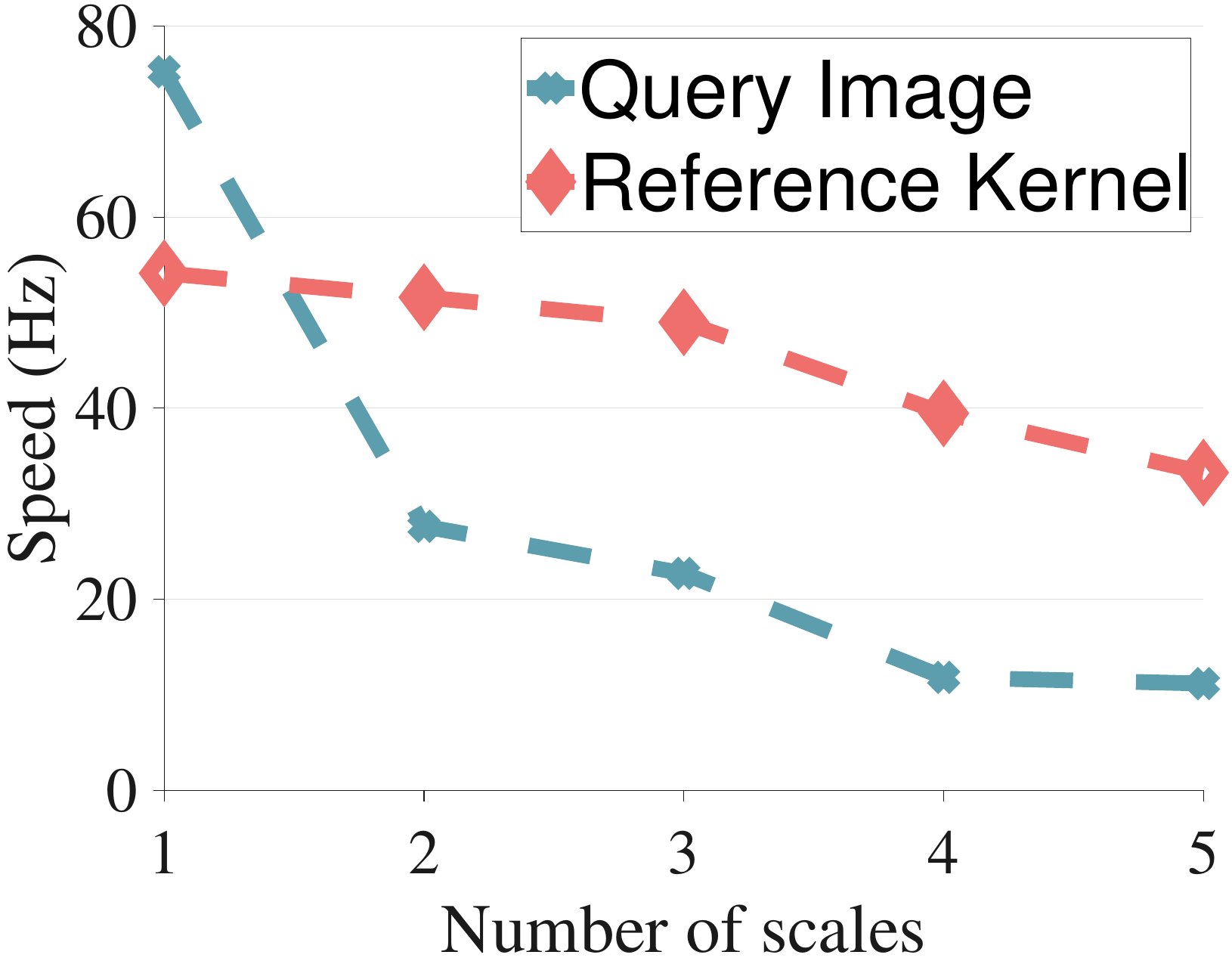}\label{fig:speed}
	}
	\caption{\textbf{Efficiency.} We compare the query image resizing presented in~\cite{liu2022gen6d} and our kernel distribution strategy from the perspectives of  multiply–accumulate operations (MACs) and speed. The number of multi-scale correlation maps $N_c$ varies from 1 to 5.}
	\label{fig:efficiency}
\end{figure}

To shed more light on the effectiveness of each component in our method, we perform ablation studies on LINEMOD from two aspects. First, we remove the three proposed modules, i.e., kernel distributing, cross-reference consistency, and decoupled estimator, from our framework. These components are then re-introduced sequentially. As shown in Table~\ref{tab:aba}, every component results in an mAP increase. Note that some alternatives exist when we estimate the object location parameters using multi-scale correlations. To highlight the effectiveness of the presented decoupled estimator, we compare it with two competitors, i.e., max pooling the $N_c$ correlation maps to a single one (max), or embedding the correlations by feeding them into convolutional layers (conv). Our decoupled estimator achieves the best performance, since the conflicting requirements w.r.t. the scale-related information are not explicitly taken into account in the competitors. Furthermore, we replace our two-stage kernel distributing approach with four different counterparts, i.e., pyramid pooling (pool), dilated convolution (dilate), kernel resizing (resize\_r), and query image resizing (resize\_q)~\cite{liu2022gen6d}. The performance consistently decreases when these methods are involved, which clearly evidences the superiority of our strategy. 

In Section~\ref{sec:method}, we claimed that our kernel distributing strategy is more computationally efficient than query image resizing~\cite{liu2022gen6d}. We thus evaluate the efficiency in terms of multiply–accumulate operations (MACs) and speed. We change the number of multi-scale correlation maps $N_c$ from 1 to 5, which accounts for different scale-related configurations. As shown in Fig.~\ref{fig:efficiency}, the MACs of the network significantly increase and the speed is much slower when multi-scale queries are involved. By contrast, the network with our kernel distributing approach is faster and requires fewer MACs in the multi-scale scenarios. This observation can be explained from two perspectives. First, estimating correlations with $N_c$ different receptive fields only requires one forward pass through the feature extraction backbone in our framework, whereas $N_c$ forward passes are needed when resizing the query image. Second, the reference kernel is much smaller than the original query image, e.g., $5\times5$ vs. $480\times640$. Consequently, the distribution process is more efficient over reference kernels. 

Demonstrated by the results in Table~\ref{tab:aba} and Fig.~\ref{fig:efficiency}, the presented kernel resizing approach is more effective and more efficient than the competitor~\cite{liu2022gen6d}.

\section{Conclusion}
\label{sec:conclusion}
In this paper, we have presented a generalizable approach to provide a location prior for previously unseen objects in a 6D object pose estimation pipeline. We have introduced a novel object translation estimator, which accounts for the conflicting requirements of object location parameters w.r.t. scale-related features. To enable multi-scale computation, we have developed a kernel distributing mechanism capable of effectively and efficiently capturing multi-scale correlations. We have conducted comprehensive experiments on LINEMOD, GenMOP, and a challenging synthetic dataset, where our method achieves state-of-the-art performance and better robustness to typical noise sources. Our findings and experimental results emphasize the importance of unseen object localization in the framework of unseen object 6D pose estimation, and we hope they will inspire further research attention in this domain.
\\

\noindent\textbf{Acknowledgments.} This work was funded in part by the Swiss National Science Foundation and the Swiss Innovation Agency (Innosuisse) via the BRIDGE Discovery grant 40B2-0\_194729.

\clearpage
{\small
\bibliographystyle{ieee_fullname}
\bibliography{egbib}
}

\end{document}